\documentclass[sigconf]{acmart}

\AtBeginDocument{%
  }

\copyrightyear{2024}
\acmYear{2024}
\setcopyright{acmlicensed}\acmConference[MM '24]{Proceedings of the 32nd
ACM International Conference on Multimedia}{28 October - 1 November 2024}{Melbourne, Australia}
\acmBooktitle{Proceedings of the 32nd ACM International Conference on
Multimedia (MM '24), 28 October - 1 November 2024, Melbourne, Australia}
\acmPrice{XXXX}
\acmDOI{XXXX}
\acmISBN{XXXX}

\acmSubmissionID{3868}

\usepackage{algorithm}
\usepackage{algorithmic}
\usepackage{bm}
\usepackage{multirow}
\usepackage{balance}
\usepackage{amsmath}
\usepackage{amsthm}
\usepackage{makecell}
\usepackage{subfigure}

\begin{document}

\title{ER-FSL: Experience Replay with Feature Subspace Learning\\for Online Continual Learning}


\author{Huiwei Lin$^\ast$}
\affiliation{%
  \institution{Harbin Institute of Technology}
  \city{Shenzhen}
  \country{China}}
\email{linhuiwei@stu.hit.edu.cn}






\thanks{$^\ast$ Corresponding Author.}

\renewcommand{\shortauthors}{Huiwei Lin.}

\begin{abstract}
  Online continual learning (OCL) involves deep neural networks retaining knowledge from old data while adapting to new data, which is accessible only once. A critical challenge in OCL is catastrophic forgetting, reflected in reduced model performance on old data. Existing replay-based methods mitigate forgetting by replaying buffered samples from old data and learning current samples of new data. In this work, we dissect existing methods and empirically discover that learning and replaying in the same feature space is not conducive to addressing the forgetting issue. Since the learned features associated with old data are readily changed by the features related to new data due to data imbalance, leading to the forgetting problem. Based on this observation, we intuitively explore learning and replaying in different feature spaces. Learning in a feature subspace is sufficient to capture novel knowledge from new data while replaying in a larger feature space provides more feature space to maintain historical knowledge from old data. To this end, we propose a novel OCL approach called experience replay with feature subspace learning (ER-FSL). Firstly, ER-FSL divides the entire feature space into multiple subspaces, with each subspace used to learn current samples. Moreover, it introduces a subspace reuse mechanism to address situations where no blank subspaces exist. Secondly, ER-FSL replays previous samples using an accumulated space comprising all learned subspaces. Extensive experiments on three datasets demonstrate the superiority of ER-FSL over various state-of-the-art methods.
\end{abstract}





\maketitle

\section{Introduction}
\label{sec:introduction}

\begin{figure*}[t]
\centering
\includegraphics[scale=0.55]{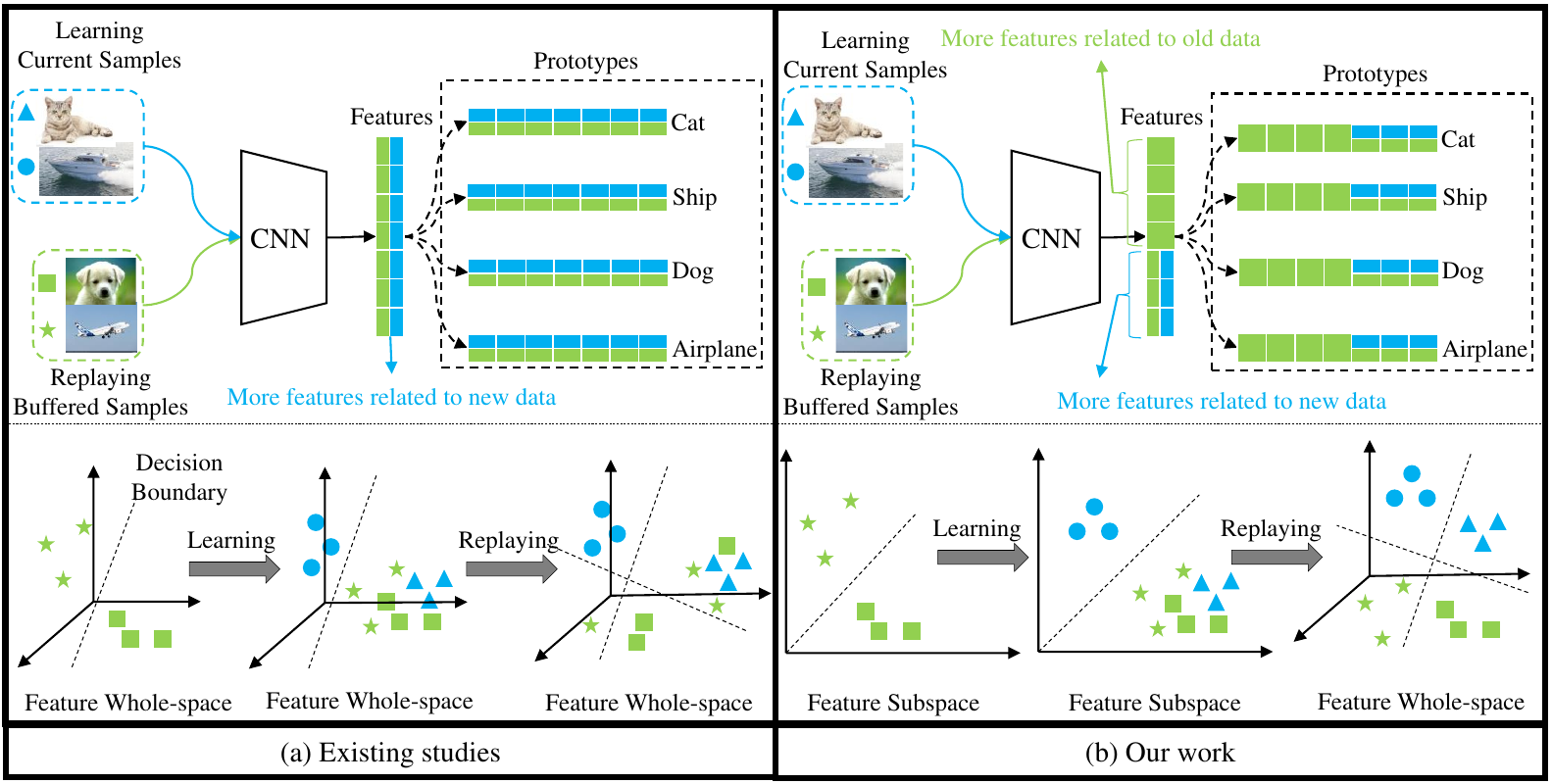}
\vspace{-3ex}
\caption{The comparison of existing studies and our work. (a) An example of existing studies. Top: the model trains all samples in the same feature space; Bottom: the change of samples in the feature space using existing methods. (b) An example of our method. Top: different from existing studies, the model learns current samples in a feature subspace while replaying buffered samples in a larger one (e.g., feature whole-space). Bottom: the change of samples in the feature space using our approach.}
\label{fig_1}%
\vspace{-2ex}
\end{figure*}

Online continual learning (OCL) is a significant problem in deep neural networks. Benefiting from offline learning on vast amounts of data, deep neural networks have demonstrated exceptional performance across various application fields, especially in the multimedia domain~\cite{xu2023rethinking,hao2021data,shen2021missing}. However, they cannot continually learn as humans do. As new data accumulates, if the model continues to use conventional training strategies, it is highly susceptible to the catastrophic forgetting (CF). This phenomenon refers to the model's performance on previously learned data significantly deteriorates after learning new data. Thus, OCL emerges as a solution to enable models to continually acquire novel knowledge from new data while retaining historical knowledge from old data. Moreover, the data can be accessed only once in an online fashion, which adds complexity to the OCL problem.

Among all methods for continual learning, replay-based methods are highly suitable for OCL to address CF problems. In this family of methods, a memory buffer is utilized to save and replay partial old data. We provide an example of OCL in a class-incremental scenario, where the model first learns classes of ``dog'' and ``airplane'', and then learns classes of ``cat'' and ``ship''. As demonstrated at the top of Figure~\ref{fig_1} (a), replay-based methods allow the model to continuously learn current samples of new data and replay buffered samples of old data. Building on this foundation, some methods have been proposed to select more important current samples for storage~\cite{aljundi2019gradient,jin2021gradient} while replaying more critical buffered samples~\cite{aljundi2019online}. At the same time, other methods~\cite{lin2023pcr} are proposed to improve the training process for more effective learning.

In this work, we dissect existing methods and empirically discover that learning and replaying in the same feature space is not conducive to addressing the forgetting issue. \emph{On one hand, training the model on new data disrupts the embedding of old data in the feature space through gradient descent}. Without replaying, the current samples occupy the main position of gradient propagation during the training process of OCL. As a result, the model learns more for correctly identifying features of new data but results in changing features of old data. It makes the old data indistinguishable in the feature space and further causes the forgetting of the model. \emph{On the other hand, although existing replay-based methods can preserve some features related to old data, the problem of changing the features of old data in the model is inevitable}. Due to the larger number of current samples compared to buffered samples, the gradient descent is still dominated by current samples. When learning and replaying occur in the same feature space, the model tends to focus more on the features of new data. For clarity, we decompose the original synchronous learning and review process. As depicted at the bottom of Figure~\ref{fig_1} (a), the samples of ``dog'' and ``airplane'' become indistinguishable in the feature space after the model has learned the samples of ``cat'' and ``ship''. Even after replaying, partial samples of ``dog'' and ``airplane'' are still indistinguishable.

With this inspiration, we intuitively explore a novel strategy to learn current samples and replay buffered samples using different feature spaces. As described at the top of Figure~\ref{fig_1} (b), the model learns current samples (blue) in a feature subspace~\cite{wang2020discriminative,jiang2022subspace,zhu2022ease} and replay buffered samples (green) in the feature whole-space. The simple yet effective strategy would ensure the model's generalization ability while improving its anti-forgetting ability. For one thing, learning current samples in a feature subspace is sufficient for the model to capture novel knowledge. For another thing, further replaying buffered samples in a larger feature space provides more features of old data for the model to retain historical knowledge. Its main idea is illustrated at the bottom of Figure~\ref{fig_1} (b). By learning in the feature subspace, the model can effectively distinguish the new data but may struggle to recognize old data. However, by replaying in the larger space, the model can retain more features of old data, thereby improving its recognition ability for old data. Consequently, the old samples that are challenging to separate in the low-dimensional space (i.e., feature subspace) can now be effectively handled in a high-dimensional space (i.e., feature whole-space), significantly alleviating the forgetting issue.

To this end, we develop a straightforward yet highly effective replay-based approach called experience replay with feature subspace learning (ER-FSL) for OCL. The fundamental motivation of ER-FSL is to employ different feature spaces for learning and replaying. Specifically, it divides the overall feature space into multiple subspaces, with each subspace used to learn a new task. Simultaneously, all learned subspaces collectively form an accumulated feature space for replaying buffered samples. This process can be mainly divided into three primary components. 1) In the \textbf{learning component}, the model utilizes the feature subspace to learn current samples and ensure its generalization ability. If there is no blank subspace, the model can take a subspace reuse mechanism to select subspaces for future task learning. 2) In the \textbf{replaying component}, buffered samples are replayed within the accumulated feature space, aiding the model in remembering more features associated with old data. 3) Based on this training way, the model can accurately identify more unknown samples within the accumulated feature space by the \textbf{testing component}.

Our main contributions can be summarized as follows:
\begin{itemize}
\item[1)] We theoretically analyze the role of feature spaces in existing methods and explore a novel strategy for utilizing separate feature spaces during learning and replaying processes. To the best of our knowledge, this work represents the first investigation into utilizing different embedding feature spaces for the replay-based OCL approaches.

\item[2)] We propose a novel OCL framework called ER-FSL to mitigate the forgetting problem by addressing the changing of old features. The primary operation involves selecting a feature subspace for learning current samples and replaying buffered samples within the accumulated feature space.

\item[3)] We conduct extensive experiments on three datasets for image classification, and the empirical results consistently demonstrate the superiority of ER-FSL over various state-of-the-art methods. We also investigate the benefits of each component by ablation studies. The source code is available at \url{https://github.com/FelixHuiweiLin/ER-FSL}.
\end{itemize}

\section{Related Work}
\label{sec:relatedwork}

\subsection{Continual Learning}
Since continual learning generally exists in various scenarios~\cite{hu2022drinking,zhang2022hierarchical,yang2022uncertainty,zhang2022continual} of deep neural networks, its related research is quite extensive. In addition to the work related to the innovation and application of continual learning methods, the analysis~\cite{pham2021continual, mirzadeh2020linear} and overview~\cite{masana2022class,mai2022online} work also have attracted much attention.

Continual learning~\cite{de2021continual,masana2022class}, also known as lifelong learning~\cite{liu2021lifelong} or incremental learning~\cite{wang2023isolation}, is a machine learning paradigm that trains models on a continuous stream of new data. It ensures the model's generalization ability~\cite{ghunaim2023real,brahma2023probabilistic} and anti-forgetting ability~\cite{dong2023heterogeneous,dong2021i3dol} at the same time. Existing methods can be generally divided into three categories. 
1) Architecture-based methods~\cite{qin2021bns,yan2021dynamically,hu2023dense} overcome CF problem by dynamical networks or static networks. dynamic networks imply that the model's network structure gradually expands with the increasing number of samples during the continual learning process, while static networks maintain their structure unchanged and allocate parameters selectively. 2) Regularization-based methods~\cite{wang2023task,pelosin2022towards,akyurek2021subspace} constrain the optimization process of the model using an additional regularization term. The design of this regularization term can be based on variations in parameters during the training process or on knowledge distillation. 3) Replay-based methods save true old data~\cite{lopez2017gradient,sun2023regularizing,luo2023class,zhou2022model} or generate pseudo old data~\cite{cui2021deepcollaboration,qi2022better} to replay with new data. And some feature replay methods~\cite{toldo2022bring} are proposed. The proposed ER-FSL in this work is a novel replay-based method.

\subsection{Online Continual Learning}
OCL is a specialized area within continual learning that emphasizes effective learning from a single pass through an online data stream, where tasks or information are introduced incrementally over time. It plays a critical role in scenarios requiring continual knowledge evolution and adaptation to new information~\cite{mai2022online}.  

Existing OCL methods are mainly based on replaying ways except AOP~\cite{guo2022adaptive}. A variety of ER-based methods have been proposed for OCL due to the effectiveness of a replay-based method called Experience replay (ER)~\cite{rolnick2019experience}. Some approaches are proposed to select more valuable samples for storing~\cite{aljundi2019gradient,jin2021gradient,he2021online} and replaying~\cite{aljundi2019online,shim2021online,wang2022improving,prabhu2020gdumb}. Other approaches~\cite{mai2021supervised,caccia2022new,zhang2022simple,guo2022online,gu2022not,lin2023pcr,chrysakis2023online,wei2023online,guo2023dealing,wang2023cba,liang2023loss,lin2023uer} belong to the model update strategy and focus on improving the training process of samples. Both SS-IL~\cite{ahn2021ss} and ER-ACE~\cite{caccia2022new} propose different cross-entropy loss functions for learning new data and reviewing old data to alleviate catastrophic forgetting. Subsequently, PCR~\cite{lin2023pcr} analyzes and integrates these two types of methods from the perspective of gradient propagation, while LODE~\cite{liang2023loss} approaches the integration from the angle of decomposing the loss function. Both of them significantly enhance the performance of the original methods. Furthermore, the performance of the latest methods~\cite{guo2022online,wei2023online} depends on multiple data augmentation operations, since data augmentation helps improve the performance of the model~\cite{zhu2021class}.

The proposed ER-FSL introduces a novel model update strategy for OCL. Different from existing strategies that learn and replay within the same feature space, the proposed ER-FSL embeds features in different spaces for current samples and buffered samples. This differentiation allows for a more effective enhancement of the model's anti-forgetting capabilities compared to existing methods. 

\section{Problem Definition and Analysis}
\label{sec:analysis}

\subsection{Problem Definition}

Taking a class-incremental scenario as an example, OCL generally considers a single-pass data stream and divides it into a sequence of $T$ learning tasks as $\mathcal{D}=\{\mathcal{D}_1,...,\mathcal{D}_T\}$, where each task $\mathcal{D}_t=\{\bm{x},y\}_{1}^{N_t}$ contains $N_t$ labeled samples. $y\in\mathcal{C}_t$ is the class label of sample $\bm{x}$, where $\mathcal{C}_t$ is the set of task-specific classes. Different tasks contain unique classes, and all of the learned classes are denoted as $\mathcal{C}_{1:t}=\bigcup_{k=1}^t \mathcal{C}_k$. The model is a neural network, consisting of a feature extractor $\bm{z}=h(\bm{x};\bm{\theta})$ and a classifier $f(\bm{z};\bm{W})=\bm{W}\cdot\bm{z}$ for the sample $\bm{x}$. $\bm{z}=[z_1,z_2,...,z_d]$ is a $d$-dimensional feature vectors, $\bm{\theta}$ and $\bm{W}=[\bm{w}_1,\bm{w}_2,...,\bm{w}_c]$ are learnable parameters, and $\bm{w}_c=[w_1^c,w_2^c,...,w_d^c]$ is a $d$-dimensional prototype vector for class $c$. OCL aims to train a unified model on data seen only once while performing well on both new and old classes.

At the beginning, the model can only access each mini-batch current samples ${\mathcal{B}\subset\mathcal{D}_t}$ once in the training process of each task. Such a training strategy is known as finetune, where the model learns without any anti-forgetting operations. Based on $\bm{z}=h(\bm{x};\bm{\theta})$, its objective loss function can be denoted as
\begin{equation}
    \label{eq_oclloss}
	L=E_{(\bm{x},y)\sim{\mathcal{B}}}[-log(\frac{exp(\bm{w}_y\cdot\bm{z})}{\sum_{c\in C_{1:t}}exp(\bm{w}_c\cdot\bm{z})})].
\end{equation}

Subsequently, a memory buffer $\mathcal{M}$ is utilized to store a small subset of observed data for replay-based methods, such as ER~\cite{rolnick2019experience}. To alleviate the forgetting problem, a mini-batch of buffered samples $\mathcal{B}_\mathcal{M}\subset\mathcal{M}$ is drawn from the memory buffer and then trained alongside current samples. Therefore, the loss function defined as Equation~(\ref{eq_oclloss}) can be improved to
\begin{equation}
    \label{eq_erloss}
	L=E_{(\bm{x},y)\sim{\mathcal{B}\cup\mathcal{B}_\mathcal{M}}}[-log(\frac{exp(\bm{w}_y\cdot\bm{z})}{\sum_{c\in C_{1:t}}exp(\bm{w}_c\cdot\bm{z})})].
\end{equation}

Finally, for each unknown sample $\bm{x}$, the model categorizes it as the class with the highest prediction probability 
\begin{equation}
 	\begin{split}
	 	y^* = \mathop{\arg\max}_{c}\frac{exp(\bm{w}_c\cdot\bm{z})}{\sum_{j\in C_{1:t}}exp(\bm{w}_j\cdot\bm{z})},c\in C_{1:t}.
 	\end{split}
\end{equation}

\begin{figure*}[t]
\centering
    \begin{minipage}[t]{0.48\linewidth}
        \centering
        \includegraphics[scale=0.19]{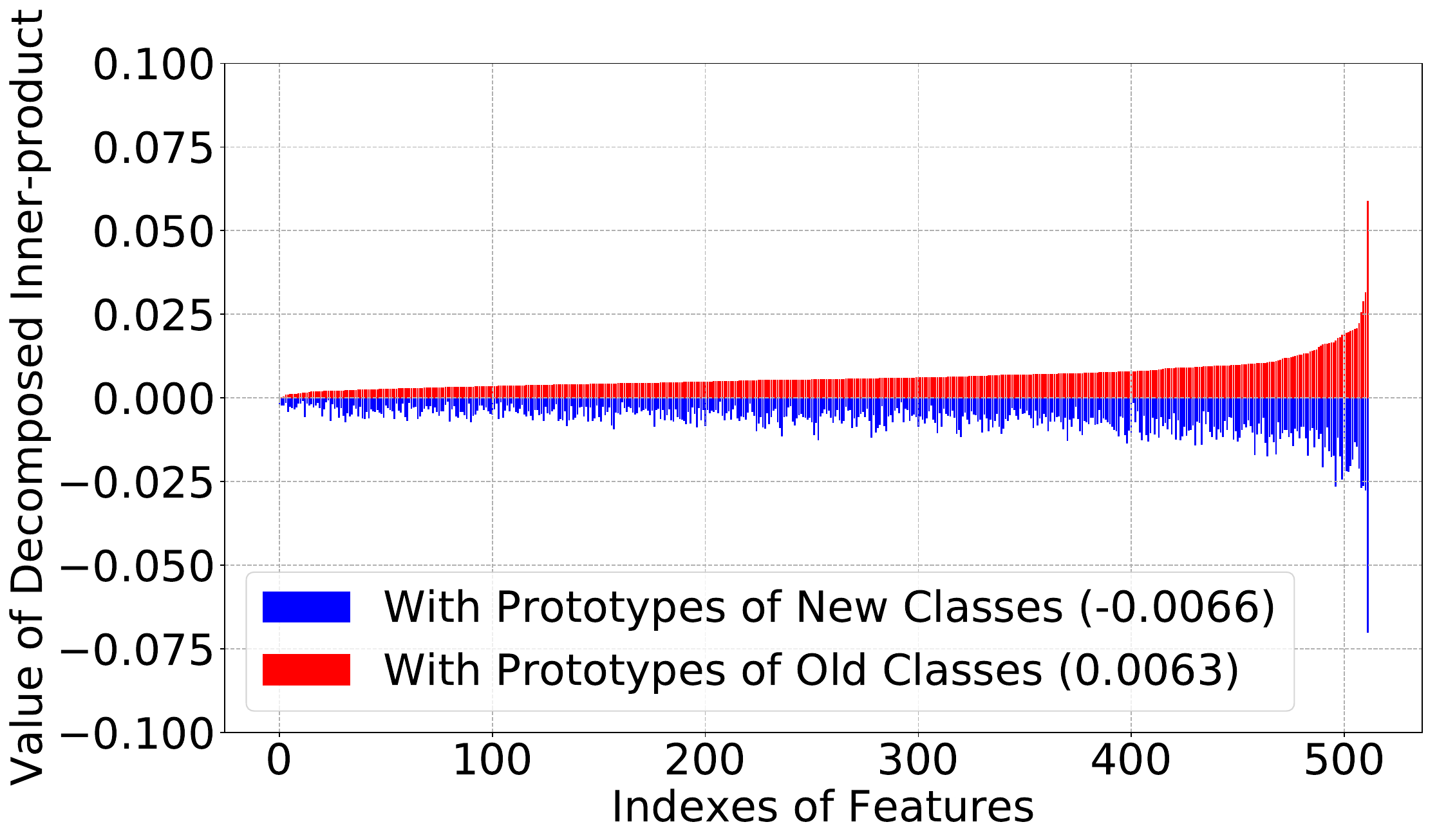}
        
        {(a) The decomposed inner-product for buffered samples after learning the first task in a general way.}
    \end{minipage}
\hspace{.1in}
    \begin{minipage}[t]{0.48\linewidth}
        \centering
        \includegraphics[scale=0.19]{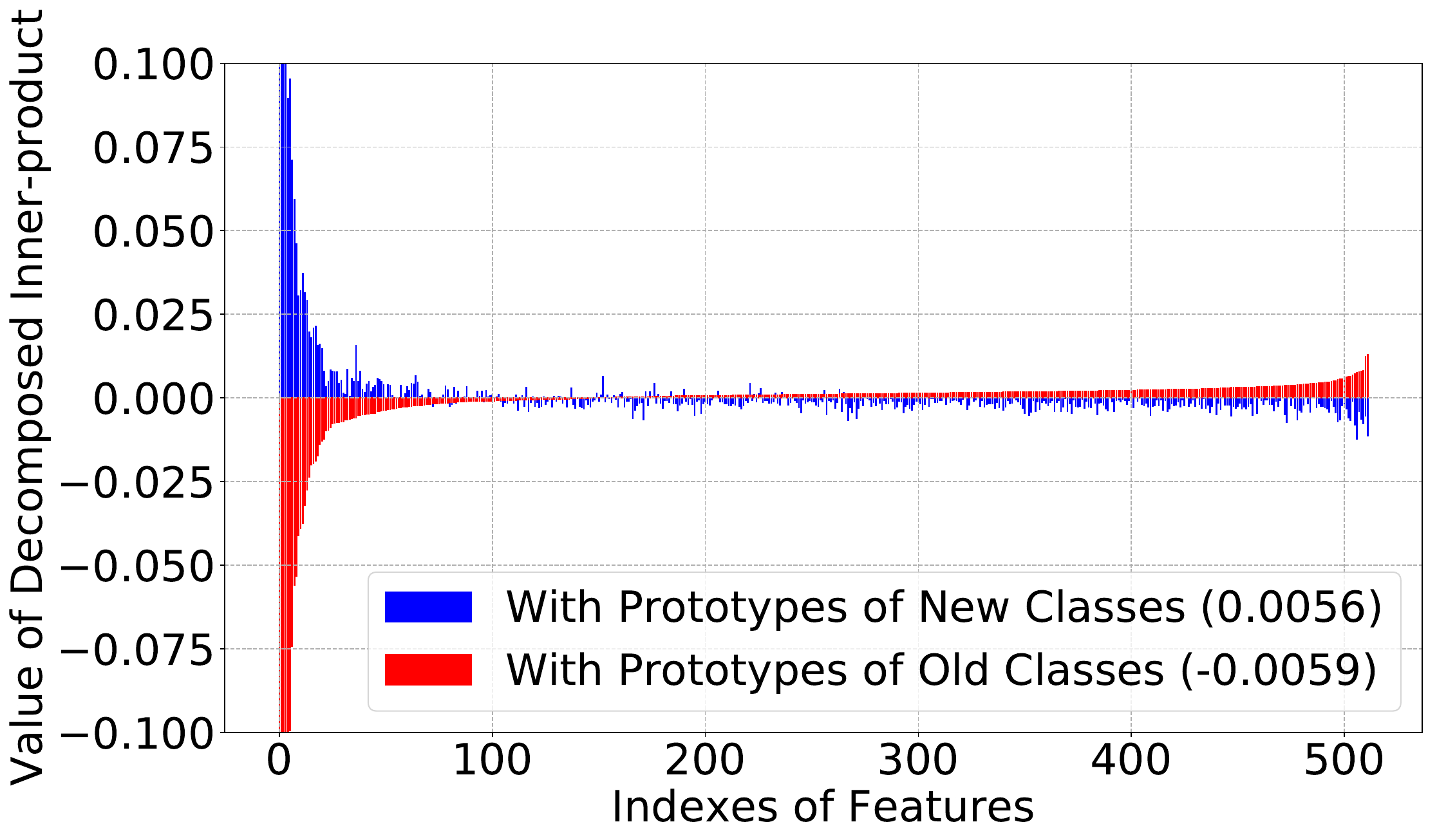}
        
        {(b) The decomposed inner-product for buffered samples after learning the second task using finetune (38.9\%).}
    \end{minipage}
    \begin{minipage}[t]{0.48\linewidth}
        \centering
        \includegraphics[scale=0.19]{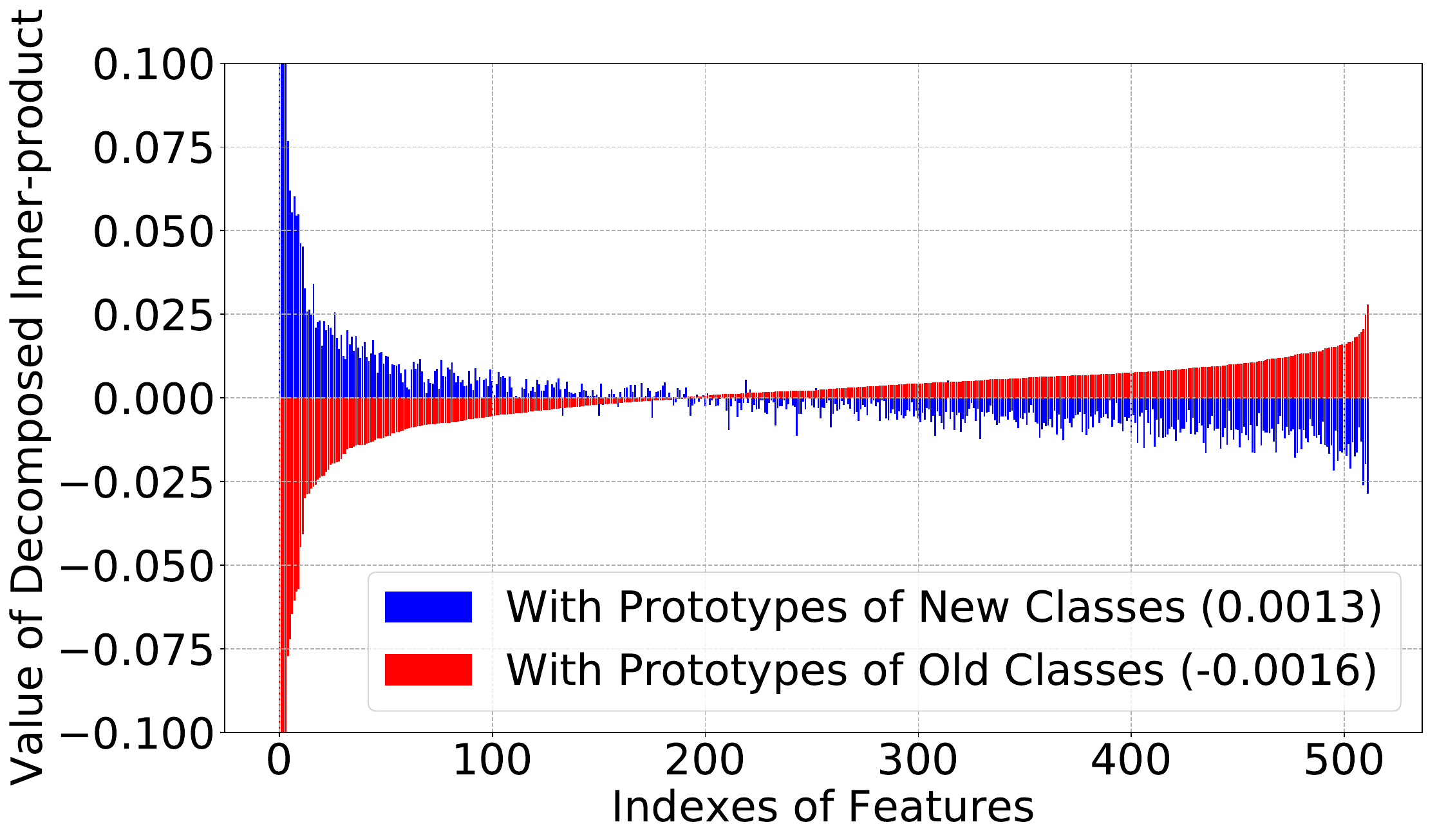}
        
        {(c) The decomposed inner-product for buffered samples after learning the second task using ER (54.2\%).}
    \end{minipage}
\hspace{.1in}
    \begin{minipage}[t]{0.48\linewidth}
        \centering
        \includegraphics[scale=0.19]{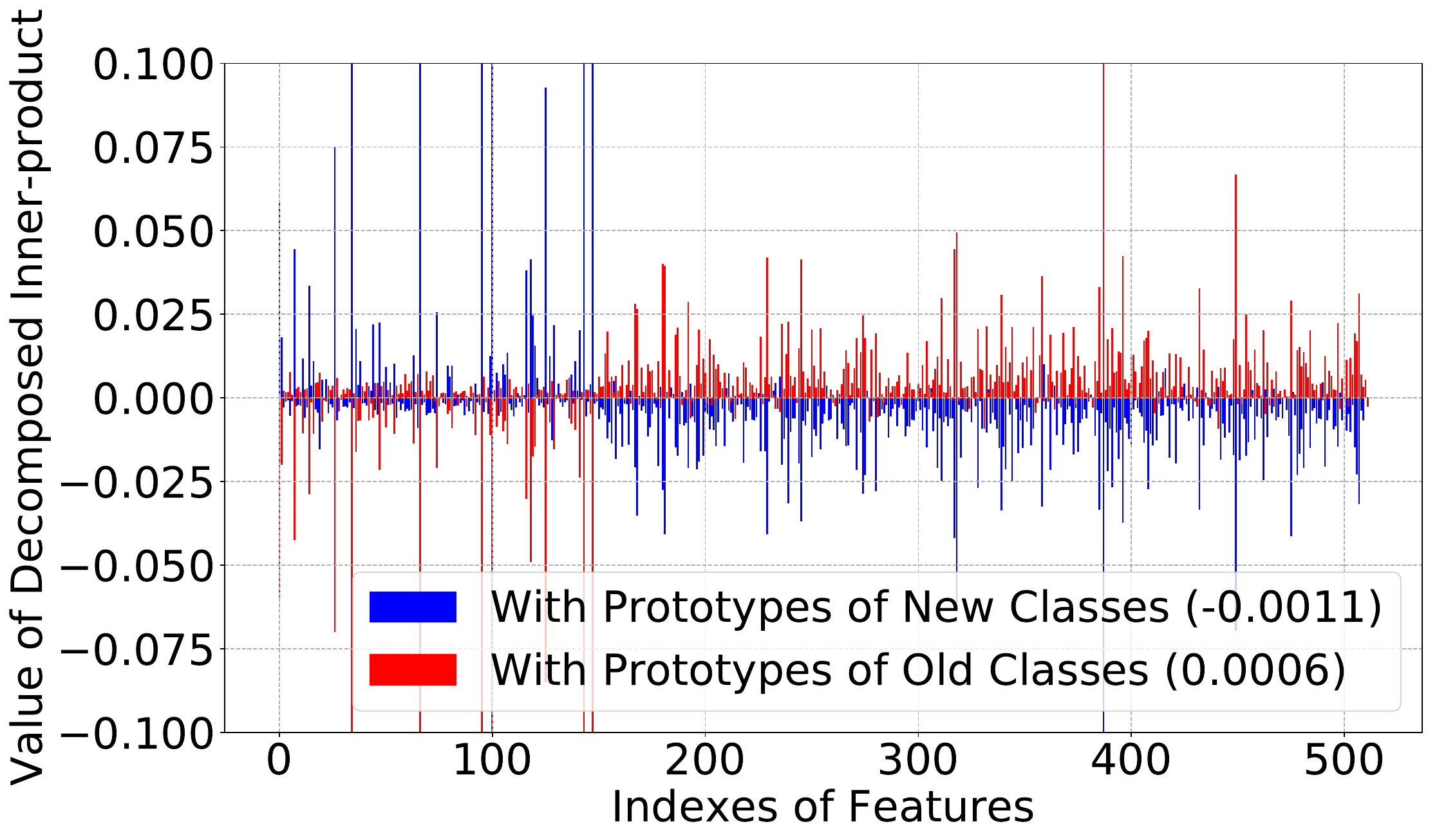}
        
        {(d) The decomposed inner-product for buffered samples after learning the second task using our way (62.4\%).}
    \end{minipage}
    \begin{minipage}[t]{0.48\linewidth}
        \centering
        \includegraphics[scale=0.19]{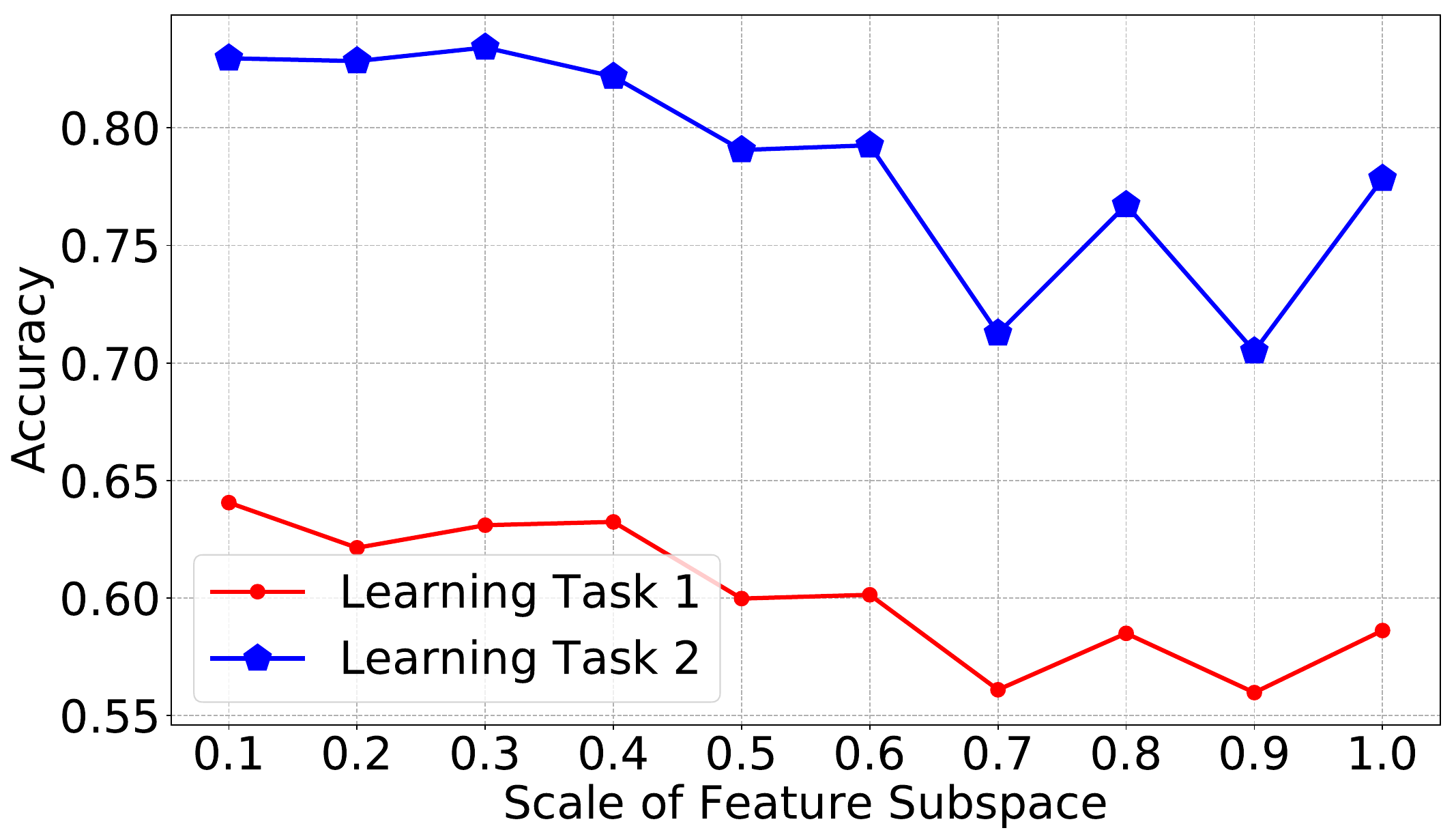}
        
        {(e) The performance in new classes of two tasks when learning in different subspaces by finetune. }
    \end{minipage}
\hspace{.1in}
    \begin{minipage}[t]{0.48\linewidth}
        \centering
        \includegraphics[scale=0.19]{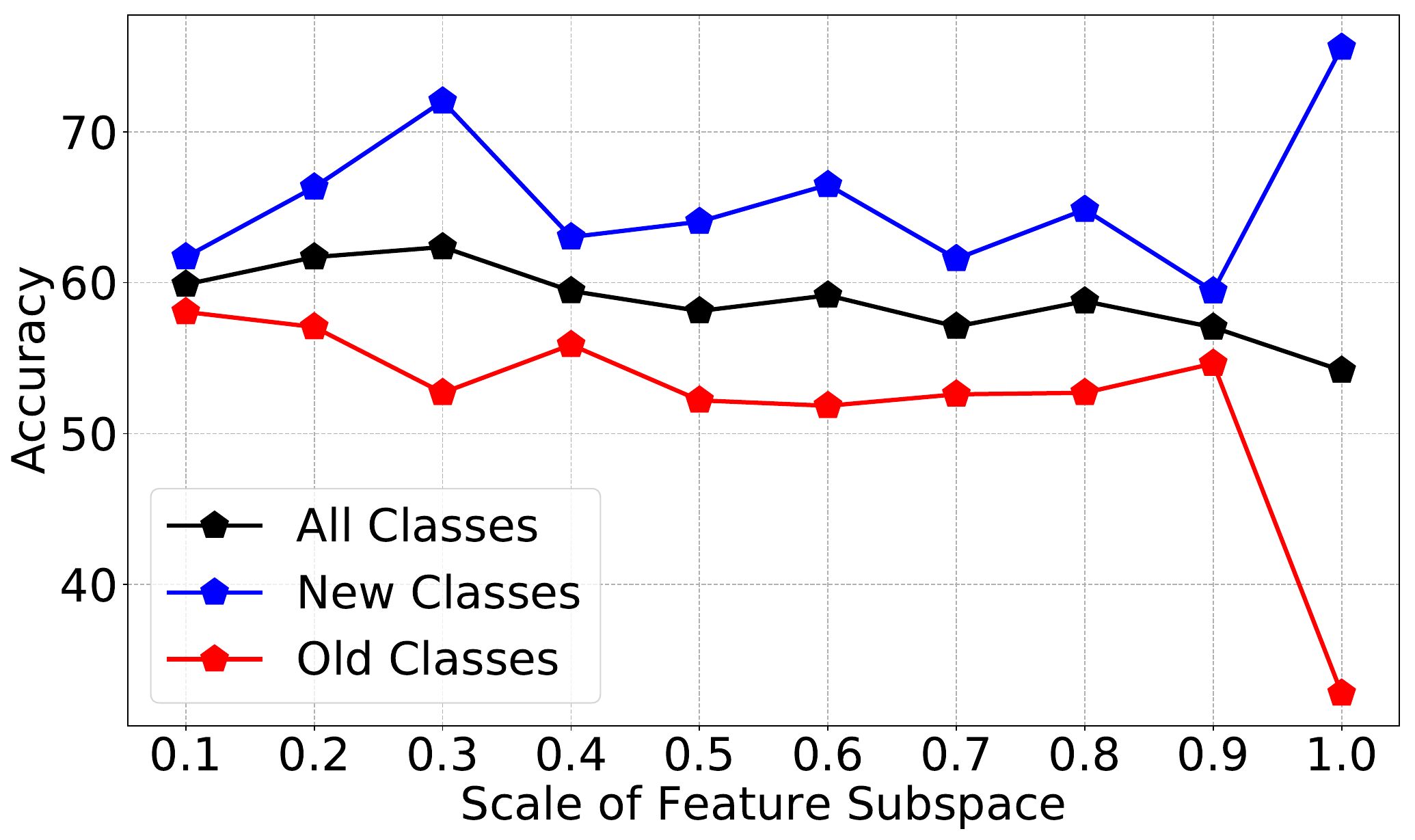}
        
        {(f) The performance on all classes when learning in different feature subspace by ER.}
    \end{minipage}

\vspace{-2ex}
\caption{The analysis results on CIFAR10 with 2 learning tasks when the memory buffer size is 1000. }
\label{fig_analysis}
\vspace{-2ex}
\end{figure*}

\subsection{Problem Exploration}
To further address the forgetting problem of the model, we analyze existing methods, explore their shortcomings, and find corresponding solutions. Specifically, we divide the CIFAR10 dataset that contains 10 classes into two tasks, each containing 5 classes, and conduct OCL analysis experiments. The model used in the experiments is Resnet18~\cite{he2016deep}, where the dimension of $\bm{z}$ is 512. All analysis results are demonstrated in Figure~\ref{fig_analysis}.

The imbalanced data, through gradient descent, results in the model focusing more on features of distinguishing new classes and ignoring features of recognizing old classes.
In the class-incremental scenario, the CF phenomenon of the model is manifested as the biased prediction $\bm{W}\cdot\bm{x}$, where the learned model tends to classify most samples into new classes. Specifically, when training a sample $\bm{x}$ of class $y$, the gradient of feature extractor can be expressed as
\begin{equation}
    \label{eq_gradient}
    \frac{\partial L}{\partial \bm{z}}=(p_y-1)\bm{w}_y+p_c\bm{w}_c,
\end{equation}
where $c\neq y$. It makes the feature $\bm{z}$ of the sample to be closer to the prototype $\bm{w}_y$ of class $y$ while keeping away from the prototypes $\bm{w}_c$ of other classes. As a result, the inner-product $\bm{w}_y\cdot\bm{z}$ will be larger than others $\bm{w}_c\cdot\bm{z}$ in the prediction. By decomposing $\bm{w}_c\cdot\bm{z}$ into $[w_1^cz_1,w_2^cz_2,...,w_d^cz_d]$, we find that the value of each dimension represents a certain feature and its importance to the sample. The greater the importance of a feature, the higher its corresponding value, and consequently, the greater its contribution to the prediction. When training the model with Equation (\ref{eq_oclloss}), all gradients are generated by the new classes. The model can only focus on features that distinguish new classes and ignore features related to old classes. Although this situation can be alleviated using Equation (\ref{eq_erloss}), the changing of features related to old classes is inevitable. Since the number of current samples is still higher than the number of buffered samples, the gradient is primarily influenced by new classes within the same feature space.

To validate this view, we calculate the decomposed inner-product between the features of buffered samples with the prototypes of old classes (red) and new classes (blue). Figure~\ref{fig_analysis} (a) shows the results after the model learning the first task. It can be seen that the values of the decomposed inner-product for old classes (red) are larger, and the model can recognize most of the buffered samples. Besides, Figure~\ref{fig_analysis} (b) and (c) show the results of completing the learning of the second task in a finetune way (Equation (\ref{eq_oclloss})) and in an ER way (Equation (\ref{eq_erloss})), respectively. If the model learns the second task using finetune, the gradient is produced by current samples from new classes. The original features (red) the model learned are changed, and the model pays more attention to the features related to new classes (blue). Hence, most of the buffered samples belonging to old classes are classified into new classes due to the biased prediction. Although existing replay-based methods such as ER can improve the values for old classes (as seen in Figure~\ref{fig_analysis} (c)), the average value for new classes (0.0013) is still higher than the one for old classes (-0.0016). Therefore, simply learning and replaying in the same feature space is not beneficial for the model to address the forgetting problem. This motivates us to investigate a new question: Why not learn current samples and replay buffered samples in different feature spaces?

\subsection{Feasibility Analysis}
\textbf{Analysis for Learning.} Learning current samples across the feature whole-space is not necessary. As illustrated in Figure~\ref{fig_analysis} (a) and (b), only a subset of the features are useful for new classes within the feature whole-space. This is due to $\bm{w}_c\cdot\bm{z}=\sum_{i=1}^{d}w^c_iz_i$, where the significance of these features will be smoothed, potentially impacting model performance, especially in larger feature spaces. To address this issue, we introduce a scaling factor to regulate the dimensions of feature and prototype vectors in a fine-tuned manner. The model's performance on new classes for each task is reported in Figure~\ref{fig_analysis} (e). The findings show that reducing the dimensions of the feature space through factors of different scales, the model's performance on new classes will be significantly improved. Meanwhile, the smaller the scale of the feature space, the smaller the fluctuation of the model.

\textbf{Analysis for Replaying.} Replaying buffered samples with a larger feature space than the one for learning can improve the ability of anti-forgetting for the model. As seen in Figure~\ref{fig_analysis} (b) and (c), due to imbalanced data, features related to old classes generally exhibit a phenomenon of weaker importance. With larger feature space, the model can memorize more features associated with old classes in the additional space, further improving the performance of old classes. We use different feature subspaces for the model to learn current samples while replaying in the feature whole-space, and the results are stated in Figure~\ref{fig_analysis} (f). The results demonstrate that with a higher dimensional feature space, the accuracy of the model on old classes has been effectively improved. As the scale changes, there is little room for improvement in this performance. This is because, for old data, too many features are not necessary either. Furthermore, we also calculate its decomposed inner-product for buffered samples when the scale is 0.3 and report the results in Figure~\ref{fig_analysis} (d). As seen in the left part, these features (indexes 1-150) are used for distinguishing new classes, since the values for new classes are larger. In the right part, these additional features (indexes 151-512), which are used to replay buffered samples, tend to correctly categorize most buffered samples as old classes.

\textbf{Summary of Analyses.} After conducting these analyses, we have synthesized our key findings as follows: (1) \emph{Learning current samples and replaying buffered samples within the same feature space is adverse to overcoming the forgetting problem.} (2) \emph{Learning current samples within a feature subspace is sufficient to ensure the generalization ability of the model.} (3) \emph{Replaying buffered samples within a larger feature space can leverage more features associated with old classes, thereby improving the anti-forgetting ability of the model.} Therefore, adopting a strategy of learning within a feature subspace while replaying within a larger feature space presents a viable approach to enhancing the model's performance.

\section{Methodology}
\label{sec:method}
Motivated by these discoveries, we develop a novel OCL framework called experience replay with feature subspace learning (ER-FSL). As stated in Figure~\ref{fig_overview}, our framework consists of a CNN-based feature extractor and a classifier. The entire workflow can be divided into the following three modules.

\subsection{Memory Buffer Module}
The setting of a memory buffer ($\mathcal{M}$) is critical for the model's performance in OCL. First, the size of the memory buffer is fixed throughout the entire training process of OCL. Second, reservoir sampling is used to screen current samples and determine whether they are stored in the memory buffer. A random sampling algorithm can extract a portion of samples from a large set and ensure that the probability of selecting each sample is equal. Third, random sampling retrieves buffered samples from the buffer for replaying.

\subsection{Continual Training Module}
The training phase of ER-FSL plays a crucial role in maintaining the model's generalization ability and anti-forgetting capability. The model can not only quickly learn novel knowledge from current samples, but also ensure the retention of historical knowledge using buffered samples as much as possible. Its objective function is

\begin{equation}
\label{eq_er_fsl}
 	\begin{split}
	 	L_{ER-FSL}=(1-\gamma)L_{c}+\gamma L_{b},
 	\end{split}
\end{equation}
where $\gamma$ is a scale factor to balance a learning component $L_{c}$ and a replaying component $L_{b}$. It encompasses a balanced optimization approach that incorporates learning novel knowledge while preserving historical knowledge.

\begin{figure}[t]
\centering
\includegraphics[scale=0.44]{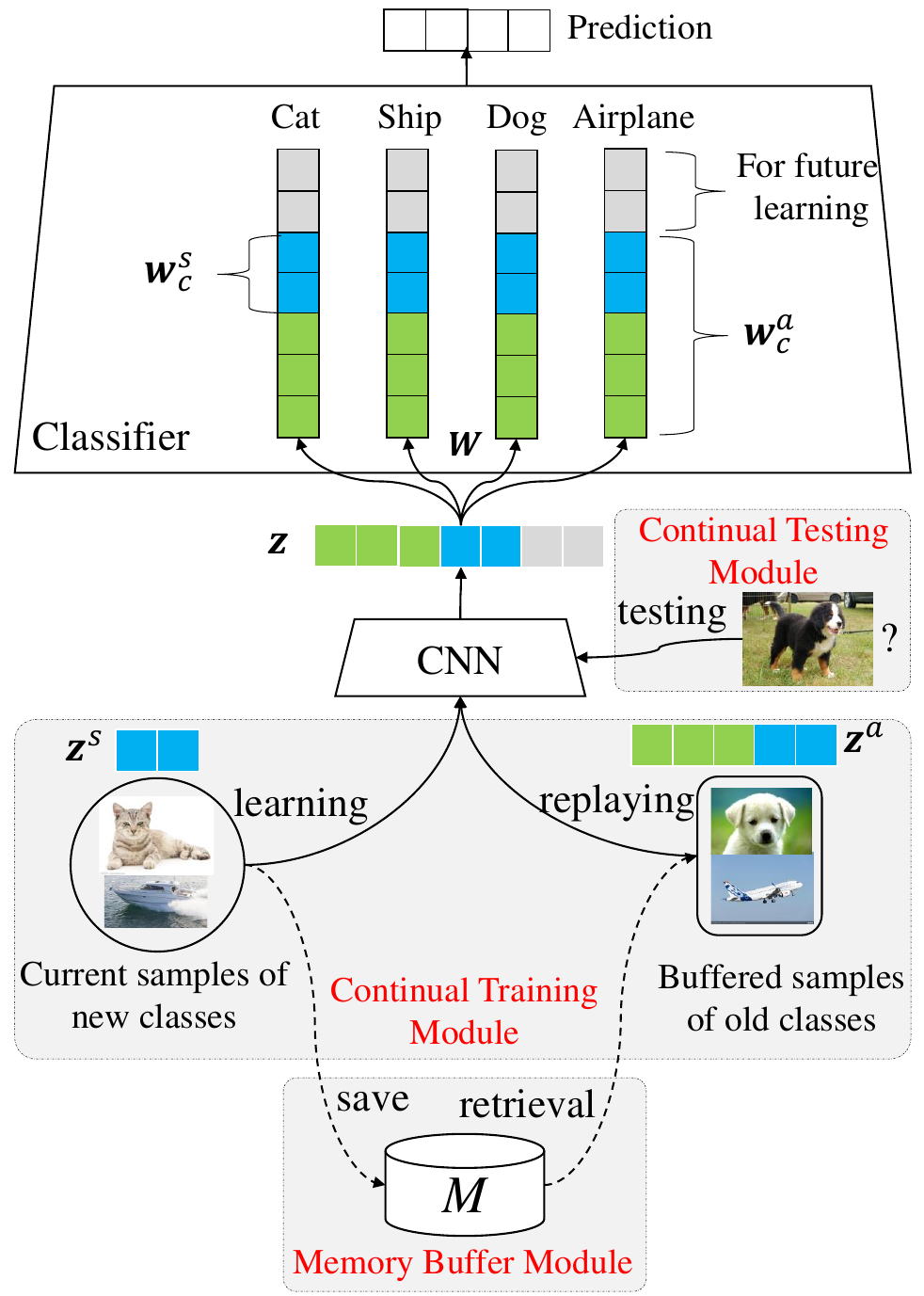}
\vspace{-3ex}
\caption{The overview of our ER-FSL framework.}
\label{fig_overview}%
\vspace{-4ex}
\end{figure}

The \textbf{learning component} $L_{c}$ is a general cross-entropy loss function only associated with current samples. It learns current samples of new classes using a feature subspace, where the novel knowledge of distinguishing new classes is saved in the subspace. Its loss function can be denoted as

\begin{equation}
    \label{eq_li}
	L_c=E_{(\bm{x},y)\sim{\mathcal{B}}}[-log(\frac{exp(\bm{w}_y^s\cdot\bm{z}^s)}{\sum_{c\in C_{1:t}}exp(\bm{w}_c^s\cdot\bm{z}^s)})].
\end{equation}
Here, $\bm{z}^s = \bm{z}\cdot\bm{S}$ is the embedding of sample $x$ and $\bm{w}_c^s = \bm{w}_c\cdot\bm{S}$ is the prototype of class $c$ in the feature subspace (the blue elements in Figure~\ref{fig_overview}). $\bm{S}$ is the diagonal matrix as  

\begin{equation}
	\bm{S}=\begin{bmatrix} 
	s_{11} & \cdots & 0\\
   \vdots &\ddots & \vdots\\
   0 & \cdots & s_{dd}
	\end{bmatrix}, s_{ii}=\left\{
    \begin{aligned}
        1&,&i\in[(t-1)k, tk) \\
        0&,&others
    \end{aligned}
    \right..
\end{equation}
It means that ER-FSL divides a $d$-dim feature space into $T$ $k$-dim subspaces, where each subspace is used to learn task $t$.

However, the overall size of the model's feature space is typically fixed, even as the number of new tasks increases. After a certain number of new tasks, the model cannot allocate a blank subspace for learning additional tasks. Hence, it is necessary to select a portion of space from the previously learned space for new tasks. 

Based on the classifier $\bm{W}\cdot\bm{z}$, the contribution of the parameters $\bm{W}$ for the $d$-th dimension is $[w_d^1z_d,w_d^2z_d,...,w_d^cz_d]$. It equals to $[w_d^1,w_d^2,...,w_d^c]\cdot z_d$, where $[w_d^1,w_d^2,...,w_d^c]$ is the $d$-th dimension of $\bm{W}$. If the variance of $[w_d^1,w_d^2,...,w_d^c]$ is larger, the features on this dimension can provide richer information to distinguish between sample classes. It means that the subspaces on dimensions with small variances contribute less and can be selected for learning new tasks. Hence, we denote the \textbf{subspace reuse mechanism} to select the subspace when there is no blank subspace as
\begin{equation}
	\bm{S}=\begin{bmatrix} 
	s_{11} & \cdots & 0\\
   \vdots &\ddots & \vdots\\
   0 & \cdots & s_{dd}
	\end{bmatrix}, s_{ii}=\left\{
    \begin{aligned}
        1&,&i\in \mathcal{K} \\
        0&,&others
    \end{aligned}
    \right..
\end{equation}    
where $\mathcal{K}$ is a subset of feature space indexes. For all elements $\mathcal{K}[j]$ ($j\in[1,k]$) in the $\mathcal{K}$, the $\mathcal{K}[j]$-th dimension of $\bm{W}$ has the top $k$ smallest variance.

The \textbf{replaying component} $L_{b}$ is also a general cross-entropy loss function only related to buffered samples. It replays buffered samples using an accumulated feature space, which consists of all learned feature space. The model uses the additional space to store features related to old data, improving the model's memory level of old knowledge. The loss function is denoted as
\begin{equation}
    \label{eq_lb}
	L_b=E_{(\bm{x},y)\sim{\mathcal{B}_\mathcal{M}}}[-log(\frac{exp(\bm{w}_y^a\cdot\bm{z}^a)}{\sum_{c\in C_{1:t}}exp(\bm{w}_c^a\cdot\bm{z}^a)})],
\end{equation}
where $\bm{z}^a = \bm{z}\cdot\bm{A}$ is the embedding of sample $x$ and $\bm{w}_c^a = \bm{w}_c\cdot\bm{A}$ is the prototype of class $c$ in the accumulated feature space (the green and blue elements in Figure~\ref{fig_overview}). $\bm{A}$ is the diagonal matrix as 
\begin{equation}
	\bm{A}=\begin{bmatrix} 
	a_{11} & \cdots & 0\\
   \vdots &\ddots & \vdots\\
   0 & \cdots & a_{dd}
	\end{bmatrix}, \label{eq_w}
    a_{ii}=\left\{
    \begin{aligned}
        1&,&i\in[0, tk) \\
        0&,&others
    \end{aligned}
    \right..
\end{equation}

\subsection{Continual Testing Module}
The \textbf{testing component} predicts testing samples by the learned model. Similar to the replaying component, each testing sample obtains its class probability distribution in the used feature whole-space. And it can be classified as

\begin{equation}
\label{eq_testting}
\hat{y}=\mathop{\arg\max}_{c}\frac{exp(\bm{w}_c^a\cdot\bm{z}^a)}{\sum_{j\in C_{1:t}}exp(\bm{w}_j^a\cdot\bm{z}^a)},c\in \mathcal{C}_{1:t}
\end{equation}

The process of this framework is described in Algorithm~\ref{alg_erfsl}. To begin with, a fixed-size memory buffer is used to save current samples (line 10) and replay previous samples (line 5). Then, the continual training module (lines 5-9) overcomes the forgetting problem by learning current samples and replaying previous samples in different feature spaces. Finally, the continual testing module (lines 14-16) predicts unknown instances by the accumulated feature space. 

\begin{figure}[t]
\centering
\includegraphics[scale=0.75]{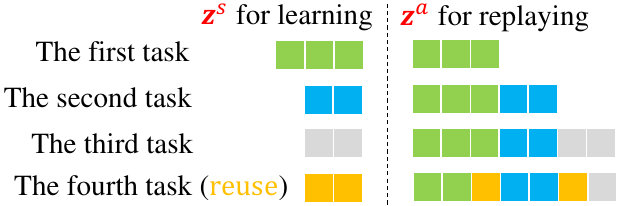}
\vspace{-3ex}
\caption{The feature space of different tasks.}
\label{fig_space}%
\vspace{-4ex}
\end{figure}

For clarity, we illustrate the subspace and accumulated space for different tasks in Figure \ref{fig_space}. For the first task, $\bm{z}^s$ and $\bm{z}^a$ are shown as the green elements. Similarly, for the second task, $\bm{z}^s$ is shown as the blue elements, and $\bm{z}^a$ is shown as the concatenation of the green and blue elements. Besides, given the third task, $\bm{z}^s$ is shown as the gray elements, and $\bm{z}^a$ is shown as the concatenation of the green, blue, and gray elements. Finally, given the fourth task, ER-FSL adopts a subspace reuse mechanism since no new subspace is available for new data. Hence, $\bm{z}^s$ is shown as the orange elements.

\begin{algorithm}[t]
\caption{ER-FSL}
\label{alg_erfsl}
\renewcommand{\algorithmicrequire}{\textbf{Input:}}
\renewcommand{\algorithmicensure}{\textbf{Output:}}
\begin{algorithmic}[1]
\REQUIRE Dataset $\mathcal{D}=\{\mathcal{D}_t\}_{t=1}^T$, Learning Rate $\lambda$, Scale $\gamma$
\ENSURE  Network Parameters $\bm{\Phi}=\{\bm{\theta},\bm{W}\}$
\STATE \textbf{Initialize}: Memory Buffer $\mathcal{M}\leftarrow\{\}$
\FOR{$\mathcal{D}_t\subset\mathcal{D}$}
\STATE //$Continual\ Training$
    \FOR{$\mathcal{B}\in \mathcal{D}_t$}
        \STATE$\mathcal{B}_\mathcal{M}\leftarrow MemoryRetrieval(\mathcal{M})$
        \STATE$L_{c}\leftarrow E_{(\bm{x},y)\sim{\mathcal{B}}}[-log(\frac{exp(\bm{w}_y^s\cdot\bm{z}^s)}{\sum_{c\in C_{1:t}}exp(\bm{w}_c^s\cdot\bm{z}^s)})]$
        \STATE$L_{b}\leftarrow E_{(\bm{x},y)\sim{\mathcal{B}_\mathcal{M}}}[-log(\frac{exp(\bm{w}_y^a\cdot\bm{z}^a)}{\sum_{c\in C_{1:t}}exp(\bm{w}_c^a\cdot\bm{z}^a)})]$
        \STATE$L\leftarrow (1-\gamma)L_{c}+\gamma L_{b}$   
        \STATE$\bm{\theta}\leftarrow\bm{\theta}+\lambda\nabla_{\bm{\theta}}L$
        \STATE$\mathcal{M}\leftarrow MemoryUpdate(\mathcal{M},\mathcal{B})$
    \ENDFOR
\STATE //$Continual\ Testing$
\STATE $m\leftarrow number\ of\ unknown\ instances$
    \FOR{$i\in\{1,2,...,m\}$}
    \STATE $\hat{y} \leftarrow \mathop{\arg\max}_{c}\frac{exp(\bm{w}_c^a\cdot\bm{z}^a)}{\sum_{j\in C_{1:t}}exp(\bm{w}_j^a\cdot\bm{z}^a)},c\in \mathcal{C}_{1:t}$
    \ENDFOR
\STATE \textbf{return} $\bm{\Phi}$
\ENDFOR
\end{algorithmic}
\end{algorithm}

\begin{table*}[t]
\renewcommand\tabcolsep{2pt}
\centering
\caption{Final Accuracy Rate (higher is better). The best scores are in boldface, and the second-best scores are underlined.}
\vspace{-2ex}
\label{tableaccuracy}
\begin{tabular}{l|cccc|cccc|cccc}
\hline
Datasets [sample size] & \multicolumn{4}{c|}{Split CIFAR10 (\%) [32$\times$32]} & \multicolumn{4}{c|}{Split CIFAR100 (\%) [32$\times$32]}& \multicolumn{4}{c}{Split MiniImageNet (\%) [84$\times$84]}\\ \hline
Buffer & \multicolumn{1}{c|}{100} & \multicolumn{1}{c|}{200} & \multicolumn{1}{c|}{500} & \multicolumn{1}{c|}{1000}    & \multicolumn{1}{c|}{500} & \multicolumn{1}{c|}{1000} & \multicolumn{1}{c|}{2000} & \multicolumn{1}{c|}{5000}     &\multicolumn{1}{c|}{500} & \multicolumn{1}{c|}{1000} & \multicolumn{1}{c|}{2000} & 5000     \\ \hline
IID          & \multicolumn{4}{c|}{55.4\scriptsize±2.5}                                  & \multicolumn{4}{c|}{17.0\scriptsize±0.9}                                     & \multicolumn{4}{c}{14.5\scriptsize±0.9}                                     \\
IID++\cite{caccia2022new}          & \multicolumn{4}{c|}{66.1\scriptsize±2.6}                                  & \multicolumn{4}{c|}{27.0\scriptsize±2.5}                                     & \multicolumn{4}{c}{21.2\scriptsize±1.8}                                     \\
FINE-TUNE          & \multicolumn{4}{c|}{16.6\scriptsize±1.4}                                  & \multicolumn{4}{c|}{5.4\scriptsize±0.5}                                     & \multicolumn{4}{c}{4.4\scriptsize±0.4}                                     \\\hline

ER (NeurIPS2019)\cite{rolnick2019experience}  & 35.5\scriptsize±2.0 & 38.8\scriptsize±3.4                 & 39.9\scriptsize±4.0                 & 43.2\scriptsize±5.6 &12.6\scriptsize±1.4& 15.7\scriptsize±1.1                  & 17.6\scriptsize±1.2                  & 16.8\scriptsize±1.5 & 10.6\scriptsize±1.0 & 12.0\scriptsize±1.2                   & 13.9\scriptsize±1.2                 & 13.9\scriptsize±2.3\\
GSS (NeurIPS2019)\cite{aljundi2019gradient} & 32.2\scriptsize±3.1& 37.1\scriptsize±3.6                 & 38.9\scriptsize±3.3                 & 43.6\scriptsize±2.9 &12.9\scriptsize±1.3& 16.1\scriptsize±0.7                  & 17.2\scriptsize±0.9                  & 17.9\scriptsize±1.2 & 10.4\scriptsize±1.1 & 12.3\scriptsize±1.0                   & 14.0\scriptsize±0.8                 & 14.6\scriptsize±0.9\\
MIR (NeurIPS2019)\cite{aljundi2019online} & 37.2\scriptsize±3.6& 41.6\scriptsize±3.9                 & 43.5\scriptsize±3.9                 & 47.7\scriptsize±4.5 &14.9\scriptsize±1.1& 17.3\scriptsize±1.6                  & 17.8\scriptsize±1.7                  & 18.4\scriptsize±1.3 & 10.9\scriptsize±0.8 & 11.5\scriptsize±0.8                   & 14.0\scriptsize±1.7                 & 14.1\scriptsize±0.8\\
ER-WA (CVPR2020)\cite{zhao2020maintaining}& 36.6\scriptsize±2.4& 39.2\scriptsize±4.4                 & 39.4\scriptsize±4.9                 & 42.9\scriptsize±4.4 &16.9\scriptsize±1.0& 19.8\scriptsize±1.3                  & 19.2\scriptsize±1.7                  & 17.8\scriptsize±1.9 & 11.2\scriptsize±1.6 & 13.4\scriptsize±1.3 & 14.5\scriptsize±0.8                   & 15.0\scriptsize±1.4                 \\
DER++ (NeurIPS2020)\cite{buzzega2020dark} & 39.1\scriptsize±3.1 & 41.9\scriptsize±3.7& 42.1\scriptsize±4.4                 & 45.7\scriptsize±3.0                 & 15.4\scriptsize±0.9 &18.0\scriptsize±1.3& 18.7\scriptsize±1.9                  & 18.7\scriptsize±1.8                  & 11.0\scriptsize±1.2 & 11.9\scriptsize±1.5                   & 12.0\scriptsize±1.8                 & 11.1\scriptsize±1.6\\
GMED (NeurIPS2021)\cite{jin2021gradient} & 34.8\scriptsize±4.1& 40.3\scriptsize±3.8                 & 42.1\scriptsize±3.5                 & 46.9\scriptsize±3.2 &14.7\scriptsize±2.9                  & 17.3\scriptsize±2.4                  & 20.7\scriptsize±2.1 & 24.1\scriptsize±2.3 & 12.1\scriptsize±1.2 & 13.1\scriptsize±1.3                   & 16.4\scriptsize±1.8                 & 17.6\scriptsize±1.7\\
ASER (AAAI2021)\cite{shim2021online} & 32.8\scriptsize±2.0& 37.5\scriptsize±3.2                 & 41.6\scriptsize±3.7                 & 40.8\scriptsize±3.7 &13.0\scriptsize±0.9& 15.9\scriptsize±1.5                  & 17.5\scriptsize±1.4                  & 18.0\scriptsize±0.9 & 9.7\scriptsize±0.7 & 12.1\scriptsize±1.3                   & 14.6\scriptsize±1.0                 & 14.5\scriptsize±2.0\\
SS-IL (ICCV2021)\cite{ahn2021ss} & 37.1\scriptsize±2.1& 42.2\scriptsize±3.3                 & 46.2\scriptsize±2.6                 & 47.6\scriptsize±2.3 &21.6\scriptsize±0.6& 23.0\scriptsize±1.3                  & 24.7\scriptsize±1.8                  & 24.9\scriptsize±1.2 & 16.7\scriptsize±1.2 & 19.3\scriptsize±1.2                   & 20.1\scriptsize±1.6                 & 23.3\scriptsize±1.2\\
SCR (CVPR-W2021)\cite{mai2021supervised} & 35.7\scriptsize±2.6& 48.5\scriptsize±1.9                 & 56.1\scriptsize±1.3                 & 57.6\scriptsize±2.2 &11.1\scriptsize±0.4& 13.9\scriptsize±0.4                  & 14.6\scriptsize±1.1                  & 15.7\scriptsize±1.0 & 10.3\scriptsize±0.7 & 12.7\scriptsize±1.2                   & 14.5\scriptsize±0.3                 & 15.9\scriptsize±0.6\\
ER-DVC (CVPR2022)\cite{gu2022not} & 32.6\scriptsize±3.3& 36.1\scriptsize±4.4                 & 37.5\scriptsize±4.2                 & 40.0\scriptsize±5.6 &14.4\scriptsize±1.4& 16.4\scriptsize±1.6                  & 18.3\scriptsize±1.3                  & 18.4\scriptsize±1.8 & 12.1\scriptsize±0.9 & 13.7\scriptsize±1.4                   & 16.0\scriptsize±1.5                 & 16.8\scriptsize±2.0\\
ER-ACE (ICLR2022)\cite{caccia2022new}   & 37.6\scriptsize±2.7& 43.6\scriptsize±2.1                 & 49.7\scriptsize±2.2                 & 50.9\scriptsize±3.0 &17.1\scriptsize±1.1& 20.8\scriptsize±1.4                  & 21.8\scriptsize±1.7                  & 23.9\scriptsize±1.4 & 13.7\scriptsize±1.1 & 15.2\scriptsize±1.3                   & 17.9\scriptsize±1.3                 & 18.3\scriptsize±1.2\\
OCM (ICML2022)\cite{guo2022online} & \textbf{48.5\scriptsize±2.2}& \textbf{53.0\scriptsize±2.3}                 & \underline{58.0\scriptsize±2.2}                 & \underline{61.3\scriptsize±2.8} &14.3\scriptsize±0.8& 17.7\scriptsize±1.4                  & 21.0\scriptsize±1.5                  & 22.7\scriptsize±0.9 & 11.8\scriptsize±0.6 & 13.3\scriptsize±1.5                   & 16.8\scriptsize±0.4                 & 18.2\scriptsize±1.0\\
OBC (ICLR2023)\cite{chrysakis2023online}   & 39.2\scriptsize±1.2& 45.1\scriptsize±2.2                 & 50.5\scriptsize±2.3                 & 51.8\scriptsize±2.3 &18.5\scriptsize±1.2& 21.5\scriptsize±0.8                  & 23.1\scriptsize±1.7                  & 23.8\scriptsize±1.6 & 12.3\scriptsize±0.6 & 14.9\scriptsize±1.5                   & 17.2\scriptsize±1.8                 & 18.3\scriptsize±1.9\\
PCR (CVPR2023)\cite{lin2023pcr}   & 40.9\scriptsize±4.1& 47.8\scriptsize±2.6                 & 52.2\scriptsize±2.9                 & 55.8\scriptsize±3.5 &\underline{21.7\scriptsize±0.9}& \underline{25.7\scriptsize±0.9}                  & \underline{27.6\scriptsize±1.4}                  & \underline{29.9\scriptsize±0.8} & \textbf{17.7\scriptsize±1.0} & \underline{19.5\scriptsize±1.5}                   & \underline{23.4\scriptsize±1.5}                 & \underline{25.0\scriptsize±2.4}\\
ER-LODE (NIPS2023)\cite{liang2023loss}   & 41.0\scriptsize±1.4& 46.3\scriptsize±1.5                 & 51.5\scriptsize±1.7                 & 53.5\scriptsize±2.4 &18.8\scriptsize±1.4& 21.6\scriptsize±0.9                  & 23.1\scriptsize±1.7                  & 24.7\scriptsize±1.7 & 15.0\scriptsize±1.3 & 16.6\scriptsize±1.1                   & 18.5\scriptsize±1.6                 & 19.5\scriptsize±1.1\\
ER-CBA (ICCV2023)\cite{wang2023cba}   & 37.9\scriptsize±2.8& 42.6\scriptsize±3.6                 & 44.3\scriptsize±2.3                 & 48.6\scriptsize±2.6 &13.5\scriptsize±0.9& 17.3\scriptsize±1.1                  & 21.3\scriptsize±1.2                  & 25.6\scriptsize±0.8 & 12.0\scriptsize±1.1 & 13.8\scriptsize±0.9                   & 16.5\scriptsize±0.4                 & 18.6\scriptsize±0.7\\
\hline
ER-FSL (Ours)   & \underline{46.9\scriptsize±2.7}& \underline{52.7\scriptsize±1.0}                 & \textbf{58.3\scriptsize±1.5}                 & \textbf{61.5\scriptsize±2.1} & \textbf{23.3\scriptsize±0.7} & \textbf{26.6\scriptsize±0.8}                  & \textbf{29.2\scriptsize±1.1}                  & \textbf{32.1\scriptsize±0.9} & \underline{17.5\scriptsize±0.1} & \textbf{21.0\scriptsize±0.9}                   & \textbf{23.6\scriptsize±1.4}                 & \textbf{27.2\scriptsize±1.2}\\
\hline
\end{tabular}
\vspace{-1ex}
\end{table*}

\begin{figure*}[t]
\centering

    \begin{minipage}[t]{0.33\linewidth}
        \centering
        \includegraphics[scale=0.17]{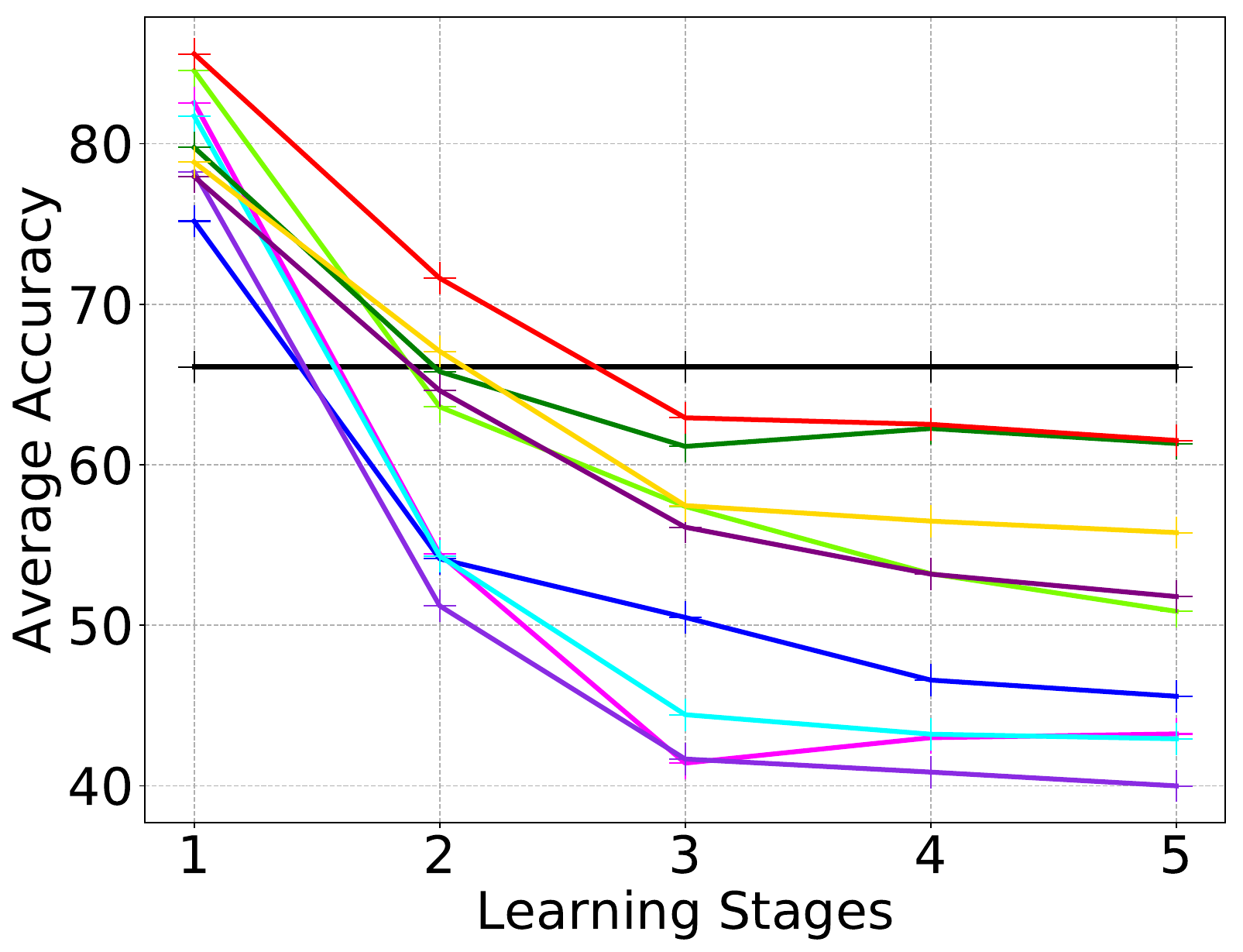}
        
        {(a) Split CIFAR10}
    \end{minipage}
    \begin{minipage}[t]{0.33\linewidth}
        \centering
        \includegraphics[scale=0.17]{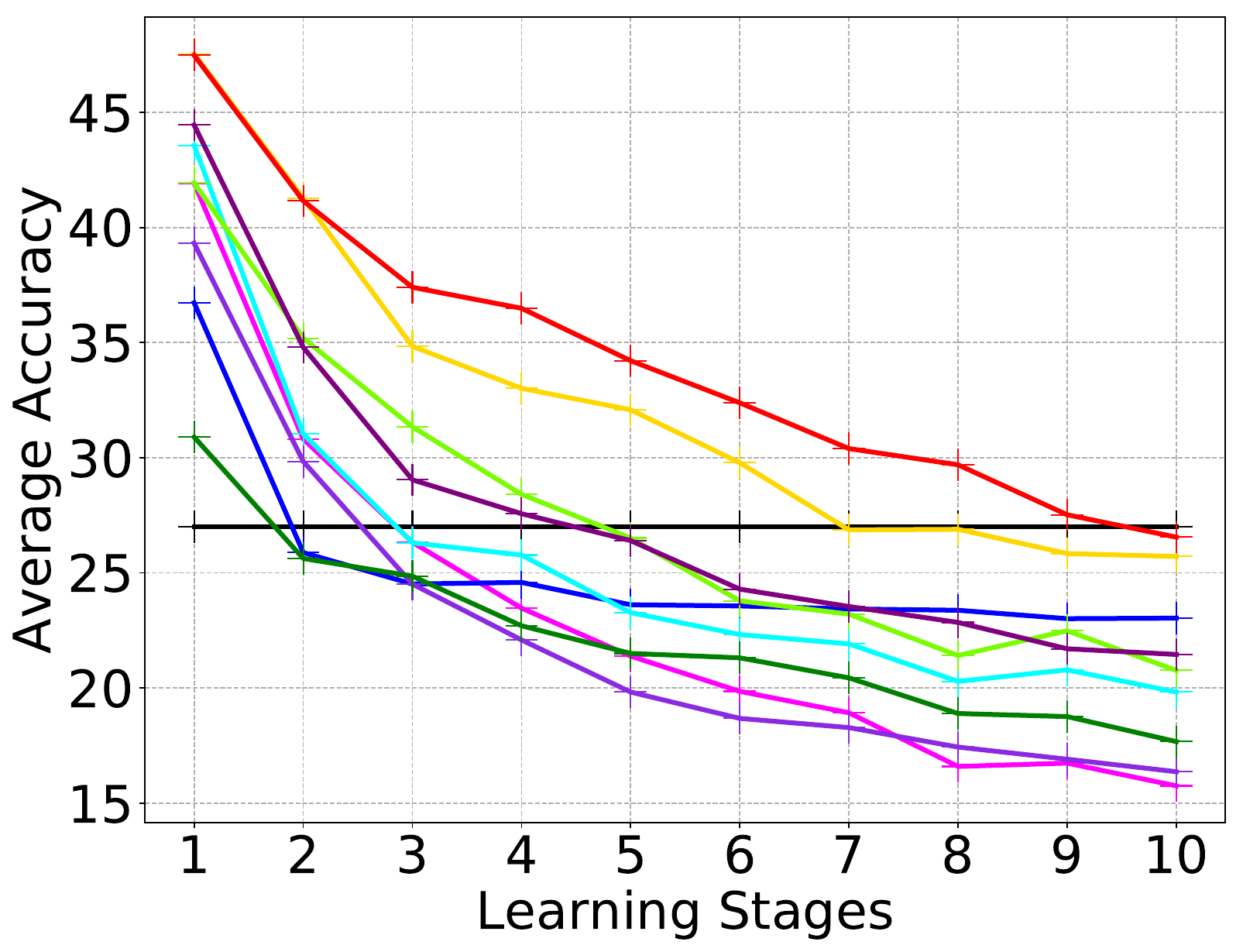}
        
        {(b) Split CIFAR100}
    \end{minipage}
    \begin{minipage}[t]{0.33\linewidth}
        \centering
        \includegraphics[scale=0.17]{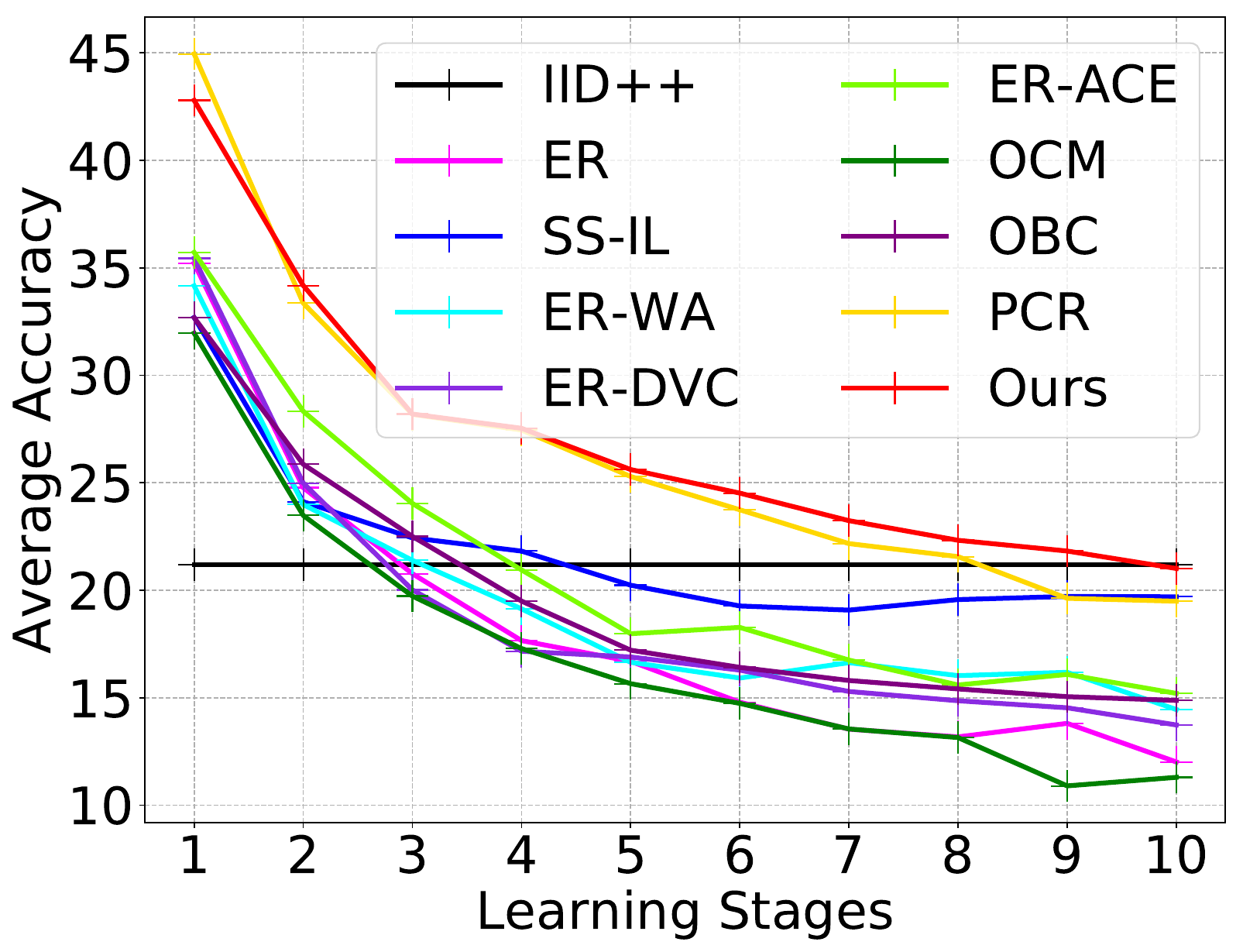}
        
        {(c) Split MiniImagenet}
    \end{minipage}
\centering
\vspace{-3ex}
\caption{Average accuracy rate on observed learning tasks on three datasets when the memory buffer size is 1000. }
\label{learningstage1000}
\vspace{-2ex}
\end{figure*}

\section{Performance Evaluation}
\label{sec:experiment}

\subsection{Evaluation Setup}
\textbf{Evaluation Datasets.} We conduct experiments on three datasets. Split CIFAR10~\cite{krizhevsky2009learning}, which is divided into 5 tasks, with each task comprising 2 classes; Split CIFAR100~\cite{krizhevsky2009learning}, and Split MiniImageNet~\cite{vinyals2016matching}, both organized into 10 tasks, each consisting of 10 classes.

\noindent\textbf{Evaluation Metrics.} Similar with~\cite{shim2021online}, we can acquire average accuracy rate $A_i$ at the $i$-th task as
\begin{equation}
\label{eq:first}
 	\begin{split}
	 	A_i = \frac{1}{i}\sum_{j=1}^{i}a_{i,j},
 	\end{split}
\end{equation}
where $a_{i,j}(j<=i)$ is the accuracy evaluated on the $j$-th task after the network has learned the first $i$ tasks. For total $T$ tasks, $A_T$ is equivalent to the final accuracy rate. 

\noindent\textbf{Implementation Details.} Similar to the recent work~\cite{wang2023cba}, we utilize ResNet18 as the feature extractor. All classes in three datasets are shuffled. The model processes 10 current samples alongside 10 buffered samples in each training step. Additionally, we employ a combination of various augmentation operations to generate the augmented samples for all methods. Hyperparameters are selected based on a validation set comprising 10\% of the training set. During the training phase, the network, initially randomly initialized, is trained using the SGD optimizer with a learning rate of 0.1.

\subsection{Overall Performance}

In this section, we conduct experiments to compare the overall performance of ER-FSL with various state-of-the-art baselines. We aim to gain insights into the strengths and weaknesses of ER-FSL.

Table~\ref{tableaccuracy} demonstrates
the final average accuracy for three datasets. All reported scores are the average score of 10 runs with a 95\% confidence interval. The results evidence that our proposed ER-FSL achieves the best overall performance. Specifically, ER-FSL achieves the best
performance under 10 of the 12 experimental scenarios. It has the most outstanding performance on Split CIFAR100 and Split MiniImageNet. For example, ER-FSL outperforms the strongest baseline PCR with a gap of 1.6\%, 0.9\%, 1.6\%, and 2.2\% on Split CIFAR100 when the size of the memory buffer is 500, 1000, 2000, and 5000, respectively. We note that ER-FSL is not optimal on Split CIFAR10 when the buffer size is smaller. Since there are fewer classes in this dataset and fewer samples in the buffer, and OCM addresses this issue using additional data augmentation operations. 

We also compare ER-FSL with CBA~\cite{wang2023cba} using the experimental setup described in CBA\cite{wang2023cba}. The learning rate is set at 0.03, which differs from the 0.1 used in our work. As shown in Table \ref{table_cba}, the results indicate that ER-FSL performs significantly better than CBA.

Figure~\ref{learningstage1000} describes the performance for some effective approaches at each task on all datasets. In the learning process, ER-FSL consistently outperforms other baselines. Especially on Split CIFAR100 and Split MiniImageNet, the performance of ER-FSL becomes increasingly evident as the number of tasks increases. For instance, ER-FSL does not surpass PCR in the initial few tasks but achieves the best in the remaining tasks as shown in Figure~\ref{learningstage1000} (b) and (c).

\begin{table}[t]
\renewcommand\tabcolsep{3pt}
\centering
\caption{Final Accuracy Rate (higher is better)/Final Forgetting Rate (lower is better ) under the setting in CBA~\cite{wang2023cba}.}
\vspace{-2ex}
\label{table_cba}
\begin{tabular}{lcccc}
\hline
Datasets & \multicolumn{2}{c}{Split CIFAR10} & \multicolumn{2}{c}{Split CIFAR100} \\
Buffer                     & 200             & 500             & 2000             & 5000            \\ \hline
CBA                 & 44.31/27.55            & 49.63/19.99            & 26.90/\textbf{9.41}             & 29.09/\textbf{8.05}            \\
ER-FSL               & \textbf{51.41}/\textbf{15.51}               & \textbf{58.43}/\textbf{12.98}               & \textbf{28.58}/9.7                & \textbf{30.14}/8.25               \\ \hline
\end{tabular}
\vspace{-2ex}
\end{table}

\begin{table}[t]
\renewcommand\tabcolsep{2pt}
\centering
\caption{Final Accuracy Rate (higher is better) for ablation study on Split CIFAR100 when the buffer size is 1000.}
\vspace{-2ex}
\label{table_ablation}
\begin{tabular}{l|lllllll}
\hline
Index &  1   &  2            &  3             & 4                          & 5                  &  6 & 7       \\ \hline
Setting     & ER-FSL & $L_c$ & $L_b$ & $\hat{y}$  &$\bm{S}$  & Inversion & $\gamma$\\ \hline
All classes &  26.6   &  17.5            &  12.6             & 7.2                          & 24.7                  &  17.8    & 24.7   \\ \hline
New classes &  33.8   &  10.5            &  38.2             & 71.6                          & 28.4                  &  18.6    & 29.5   \\ \hline
Old classes & 25.7    &  18.2            &  9.7             & 0.0                          & 24.2                  &  17.8      & 24.2  \\ \hline
\end{tabular}
\vspace{-3ex}
\end{table}

\begin{figure}[t]
\centering
\includegraphics[scale=0.16]{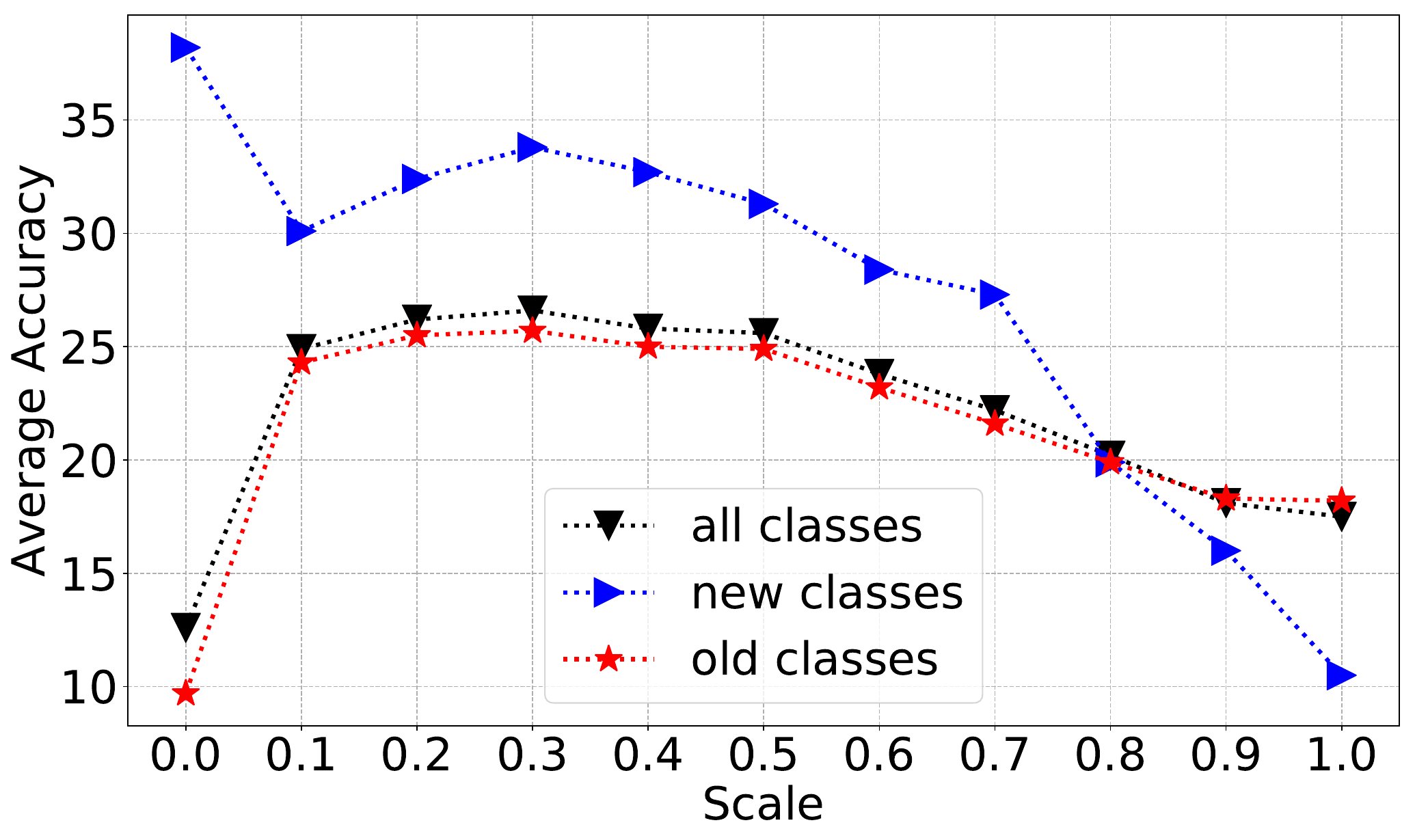}
\vspace{-3ex}
\caption{The performance of ER-FSL on Split CIFAR100 (buffer size=1000) with different values of $\gamma$.}
\label{fig:scale}%
\vspace{-2ex}
\end{figure}


\subsection{Ablation Study}

We conduct ablation experiments to analyze the contribution of various components and choices made in ER-FSL, and the results are stated in Table~\ref{table_ablation}. First, the learning component $L_c$ is necessary to ensure the model's generalization ability. Without it (Index 2), the model's overall performance is significantly decreased. Besides, the replaying component $L_b$ overcomes the forgetting issue, as the accuracy is low when the replaying component is removed (Index 3). Finally, if the model uses the same subspace as $\bm{z}^s$ when testing, the model can not memorize anything of old data (Index 4). Therefore, the model should predict unknown instances in the space as $\bm{z}^a$. 

In the meantime, some subtle settings are also important. First, it is necessary to assign different feature subspaces to each task. If we use a fixed $S$ to select the same subspace, the performance of the model will decrease (Index 5). Second, the spaces used for replaying and learning cannot be inversed. Since the inversion version (Index 6, replaying in a feature subspace and learning in a larger space) performs worse than the original one (Index 1). Third, the scale $\gamma$ is vital to balance the novel and historical knowledge. Without it, the Equation (\ref{eq_er_fsl}) becomes $L_c+L_b$, which can not achieve the best results (Index 7). Moreover, we report the performance of the model with different $\gamma$ in Figure~\ref{fig:scale}. The results indicate that a suitable $\gamma$ can enable the model to perform better on both old and new data.

Furthermore, we even conduct experiments on Split CIFAR100 with different subspace sizes and report the results in Table~\ref{table:vary}. Firstly, the performance of ER-FSL improves as the subspace size increases since a larger space can capture more useful features. Secondly, the results show that the best subspace size is 100, which triggers the subspace reuse mechanism. It means that the results of ER-FSL in Table~\ref{tableaccuracy} could be better; however, for a fair comparison with other methods, we have to limit our method to choosing a feature subspace of size 51 to fill the entire feature space (size = 512). Thirdly, although the performance tends to decrease when the size further increases, the decline is not significant.

\subsection{Complexity Analysis}
The final comparison between the existing methods and ER-FSL regarding computation and memory complexity is illustrated in Table \ref{table:time}. We compare the computation complexity using $C_\mathcal{D}$~\cite{ghunaim2023real}, which is determined by the relative training FLOPs. For example, we set the computation complexity of ER as 1. Since ER-FSL, ER-ACE, and PCR straightly modify the loss of ER, their computational complexities are equivalent to 1. However, OCM heavily relies on massive data augmentation operations, making its relative complexity 16. Meanwhile, we set the relative memory as the memory complexity. For instance, the memory complexity of ER is set as 1. OCM has a complexity of 2 due to the need to save an additional model for knowledge distillation, while OBC uses two classifiers, making its complexity greater than 1. The memory complexity of ER-FSL is 1. Thus, ER-FSL is also excellent in terms of complexity.

\begin{table}[t]
\renewcommand\tabcolsep{1pt}
\centering
\vspace{-2ex}
\caption{The performance of ER-FSL on Split CIFAR100 with different sizes of subspaces (buffer size=1000).}
\vspace{-2ex}
\label{table:vary}
\begin{tabular}{l|ccccc|c|cccc}
\hline
Subspace size     & 10 & 20 & 30 & 40 & 50 & 51 (our)& 100 & 200  & 300 & 400  \\ \hline
Subspace reuse          & \multicolumn{5}{c|}{no} & no & \multicolumn{4}{c}{yes}  \\
\hline
Final accuracy          & 25.3  & 26.1     & 26.9 & 27.0  & 26.9 & 26.6 & 27.6 & 27.0  & 25.9 & 24.8 \\
\hline

\end{tabular}
\vspace{-3ex}
\end{table}

\begin{table}[t]
\renewcommand\tabcolsep{2pt}
\centering
\caption{The computation and memory complexity.}
\vspace{-2ex}
\label{table:time}
\begin{tabular}{l|ccccccc}
\hline
Metric       & ER      & ER-ACE & OBC & OCM & PCR &Ours\\ \hline
Computation ($C_\mathcal{D}$)           & 1      & 1 & 1.5 & 16 & 1 & 1 \\
\hline
Memory (Model)           & 1      & 1 & >1 & 2 & 1 &1 \\
\hline
\end{tabular}
\vspace{-4ex}
\end{table}
\section{Conclusion}
\label{sec:conclusion}
In this paper, we develop a simple yet effective OCL method called ER-FSL to alleviate the CF. By examining the change of features, we find that learning and replaying in the same feature space is not beneficial for the anti-forgetting of the model. Based on this observation, a novel ER-FSL is proposed to learn and replay in different feature spaces for OCL. It divides the entire feature space into multiple feature subspaces, where each subspace is used to learn each new task. Meanwhile, it replays previous samples using an accumulated feature space, which consists of all learned feature subspaces. Extensive experiments on three datasets demonstrate the superiority of ER-FSL over various state-of-the-art baselines.



\bibliographystyle{ACM-Reference-Format}
\balance





\end{document}